\documentclass[10pt,twocolumn,twoside] {IEEEtran}
\def\comment#1{}
\usepackage{graphicx,subfigure,subfig,epsfig,amsfonts,amsmath,amssymb,lscape,color,url}
\usepackage[noadjust]{cite}
\DeclareSymbolFont{grb}{OML}{cmm}{b}{it}
\DeclareMathSymbol{\alphab}{\mathord}{grb}{"0B}
\DeclareMathSymbol{\betab}{\mathord}{grb}{"0C}
\DeclareMathSymbol{\gammab}{\mathord}{grb}{"0D}
\DeclareMathSymbol{\deltab}{\mathord}{grb}{"0E}
\DeclareMathSymbol{\epsilonb}{\mathord}{grb}{"0F}
\DeclareMathSymbol{\zetab}{\mathord}{grb}{"10}
\DeclareMathSymbol{\etab}{\mathord}{grb}{"11}
\DeclareMathSymbol{\thetab}{\mathord}{grb}{"12}
\DeclareMathSymbol{\kappab}{\mathord}{grb}{"14}
\DeclareMathSymbol{\lambdab}{\mathord}{grb}{"15}
\DeclareMathSymbol{\mub}{\mathord}{grb}{"16}
\DeclareMathSymbol{\nub}{\mathord}{grb}{"17}
\DeclareMathSymbol{\rhob}{\mathord}{grb}{"1A}
\DeclareMathSymbol{\sigmab}{\mathord}{grb}{"1B}
\DeclareMathSymbol{\taub}{\mathord}{grb}{"1C}
\DeclareMathSymbol{\phib}{\mathord}{grb}{"1E}
\DeclareMathSymbol{\psib}{\mathord}{grb}{"20}
\DeclareMathSymbol{\omegab}{\mathord}{grb}{"21}
\DeclareMathSymbol{\epsilonb}{\mathord}{grb}{"22}
\DeclareMathSymbol{\varphib}{\mathord}{grb}{"27}
\DeclareMathSymbol{\chib}{\mathord}{grb}{"1F}
\def\comment#1{}

\newcommand{\fv}{\boldsymbol{f}}

\newcommand{\nv}{\boldsymbol{n}}

\newcommand{\xv}{\boldsymbol{x}}
\newcommand{\yv}{\boldsymbol{y}}

\begin{document}
\setlength{\parskip}{.02in}

\title{Multiscale Shrinkage and L\'{e}vy Processes}

\author{
\authorblockN{Xin Yuan, Vinayak Rao, Shaobo Han and Lawrence Carin} \\
\authorblockA{Department of Electrical and Computer Engineering\\
Duke University \\
Durham, NC 27708-0291 USA }
}

\maketitle

\begin{abstract}
A new shrinkage-based construction is developed for a compressible vector $\boldsymbol{x}\in\mathbb{R}^n$, for cases in which the components of $\xv$ are naturally associated with a tree structure. Important examples are when $\xv$ corresponds to the coefficients of a wavelet or block-DCT representation of data. The method we consider in detail, and for which numerical results are presented, is based on increments of a gamma process. However, we demonstrate that the general framework is appropriate for many other types of shrinkage priors, all within the L\'{e}vy process family, with the gamma process a special case. Bayesian inference is carried out by approximating the posterior with samples from an MCMC algorithm, as well as by constructing a
heuristic variational approximation to the posterior. We also consider expectation-maximization (EM) for a MAP (point) solution. State-of-the-art results are manifested for compressive sensing and denoising applications, the latter with spiky (non-Gaussian) noise.
\end{abstract}

\section{Introduction\label{Intro}}

In signal processing, machine learning and statistics, a key aspect of modeling concerns the compact representation of data. Consider an $n$-dimensional signal $\boldsymbol{T}\xv$, where $\boldsymbol{T}\in\mathbb{R}^{n\times n}$ and $\xv\in\mathbb{R}^n$; the columns of $\boldsymbol{T}$ define a basis for representation of the signal, and $\xv$ are the basis coefficients. Two important means of constituting the columns of $\boldsymbol T$ are via orthornormal wavelets or the block discrete cosine transform (DCT) \cite{Mallat}. The former is used in the JPEG2000 compression standard, and the latter in the JPEG standard; compression is manifested because many of the components of $\xv$ may be discarded without significant impact on the accuracy of the reconstructed signal \cite{Shapiro93SPT,Xiong96SPL}. Wavelet and block-DCT-based compression are applicable to a wide class of natural data. For such data, the original $\xv$ typically has no values that are exactly zero, but many that are relatively small, and hence in this case $\xv$ is termed ``compressible."

In compression, one typically acquires the uncompressed data, sends it through filters defined by $\boldsymbol T$, and then performs compression on the resulting $\xv$. In inverse problems, one is not given $\xv$, and the goal is to recover it from given data. For example, in compressive sensing (CS) \cite{Candes05ITT,Donoho06ITT,Candes06ITT} one is typically given $m$ linear projections of the data, $\yv=\boldsymbol{HT}\xv+\nv$, where $\boldsymbol{H}\in\mathbb{R}^{m\times n}$ and $\nv\in\mathbb{R}^m$ is measurement noise (ideally $m\ll n$). The goal is to recover $\xv$ from $\yv$. In denoising applications \cite{Portilla03IPT}
$\boldsymbol{H}$
may be the $n\times n$ identity matrix, and the goal is to recover $\xv$ in the presence of noise $\nv$, where the noise may be non-Gaussian. If one is performing recovery of missing pixels \cite{Zhou11AISTATS} (the ``inpainting'' problem), then $\boldsymbol{H}$ is defined by $m$ rows of the $n\times n$ identity matrix. Other related problems include deblurring, where $\boldsymbol{H}$ may be a blur kernel \cite{blur}.

To solve these types of problems, it is important to impart prior knowledge about $\xv$, where compressibility is widely employed. As a surrogate for compressibility, there is much work on assuming that $\xv$ is exactly sparse \cite{Candes05ITT,Donoho06ITT,Candes06ITT}; the residuals associated with ignoring the small components of $\xv$ are absorbed in $\nv$. This problem has been considered from a large range of perspectives, using methods from optimization \cite{Tropp04ITT,Tropp07ITT,Baraniuk10ITT} as well as Bayesian approaches \cite{Ji08SPT,He09SPT,Baron10SPT}. In this paper we take a Bayesian perspective. However, it is demonstrated that many of the shrinkage priors we develop from that perspective have a corresponding regularizer from an optimization standpoint, particularly when considering a maximum \emph{a posterior} (MAP) solution.

In Bayesian analysis researchers have developed methods like the relevance vector machine (RVM) \cite{Tipping01RVM}, which has been applied successfully in CS \cite{Ji08SPT}. Spike-slab priors have also been considered, in which exact sparsity is imposed \cite{He09SPT}. Methods have also been developed to leverage covariates ($e.g.$, the spatial locations of pixels in an image \cite{Zhou11AISTATS}), using methods based on kernel extensions. Inference has been performed efficiently using techniques like Markov chain Monte Carlo (MCMC) \cite{Zhou11AISTATS}, variational Bayes \cite{He10SPL}, belief propagation \cite{Baron10SPT}, and expectation propagation \cite{Seeger}. Expectation-maximization (EM) has been considered \cite{zhang2012ep} from the MAP perspective.

Recently there has been significant interest on placing ``global-local shrinkage'' priors \cite{Polson10shrinkglobally,Polso12Levy} on the components of $\xv$, the $i$th of which is denoted $x_i$. Each of these priors imposes the general construction $x_i\sim\mathcal{N}(0,\tau^{-1}\alpha_i^{-1})$, with a prior on
$\alpha_i^{-1}$ highly concentrated at the origin, with heavy tails. The set of ``local'' parameters $\{\alpha_i\}$, via the heavy tailed prior, dictate which components of $\xv$ have significant amplitude. Parameter $\tau$ is a ``global'' parameter, that scales $\{\alpha_i\}$ such that they are of the appropriate amplitude to represent the observed data. Recent examples of such priors include the horseshoe \cite{Carvalho_horseshoe} and three-parameter-beta priors \cite{Armagan12}.

In this line of work the key to imposition of compressibility is the heavy-tailed prior on the
$\{x_i\}$. An interesting connection has been made recently between such a prior and increments of general L\'{e}vy processes \cite{Polso12Levy}. It has been demonstrated that nearly all of the above shrinkage priors may be viewed in terms of increments of a L\'{e}vy process, with appropriate L\'{e}vy measure. Further, this same L\'{e}vy perspective may be used to manifest the widely employed spike-slab prior. Hence, the L\'{e}vy perspective is a valuable framework from which to view almost all priors placed on $\xv$. Moreover, from the perspective of optimization, many regularizers on $\xv$ have close linkages to the L\'{e}vy process 
\cite{Zhang12NIPS}.

All of the shrinkage priors discussed above assume that the elements of $\xv$ are exchangeable; $i.e.$, the elements of $\{\alpha_i\}$ are drawn i.i.d. from a heavy-tailed distribution. However, the components of $\xv$ are often characterized by a tree structure, particularly in the context of a wavelet \cite{Shapiro93SPT,He09SPT} or block-DCT \cite{Xiong96SPL,He10SPL} representation of $\boldsymbol T$. It has been recognized that when solving an inverse problem for $\xv$, the leveraging of this known tree structure may often manifest significantly improved results 
\cite{BaraniukMCS}.

The principal contribution of this paper concerns the extension of the aforementioned shrinkage priors into a new framework, that exploits the known tree structure. We primarily focus on one class of such shrinkage priors, which has close connections from an optimization perspective to adaptive Lasso \cite{ZouAdaLasso}. However, we also demonstrate that this specific framework is readily generalized to all of the shrinkage priors that may be represented from the L\'{e}vy perspective. This paper therefore extends the very broad class of L\'{e}vy-based shrinkage priors, leveraging the known tree structure associated with $\boldsymbol T$.

We principally take a Bayesian perspective, and perform computations using MCMC and variational Bayesian analysis. However, we also perform a MAP-based point estimation, which demonstrates that this general tree-based structure may also be extended to an optimization perspective. Specific applications considered are estimation of $\xv$ in the context of CS and in denoising. For the latter we consider non-Gaussian noise,
specifically noise characterized by the sum of Gaussian and spiky components. This is related to the recent interest in robust principal component analysis \cite{Candes11RPCA}. We demonstrate that this noise may also be viewed from the perspective of a L\'{e}vy jump process. State-of-the-art results are realized for these problems.

The remainder of the paper is organized as follows. The basic setup of the inverse problems considered is detailed in Section \ref{sec:model}. In that section we also introduce the specific class of global-local priors for which example results are computed. In Section \ref{sec:TreeShrink} we discuss the method by which we extend such a shrinkage prior to respect the tree structure of $\xv$, when associated with expansions $\boldsymbol T$ tied to wavelet, block-DCT and other related bases. It is demonstrated in Section \ref{sec:Levy} how the above model is a special case of a general class of tree-based models, based on the L\'{e}vy process. As indicated above, this is an active area of research, and in Section \ref{sec:related} we provide a detailed discussion of how the current work is related to the literature. In Section \ref{Sec:Spiky} we discuss how spiky noise is modeled, and its connection to the L\'{e}vy process. Three different means of performing inference are discussed in Section \ref{sec:inference}, with further details in the Appendix. An extensive set of numerical results and comparisons to other methods are presented in Section \ref{Sec:Result}, with conclusions discussed in Section \ref{sec:conclusions}.

\section{Tree-Based Representation and Coefficient Model\label{sec:model}}

\subsection{Measurement model}
\label{Sec:model_notree}
Consider the wavelet transform of a two-dimensional (2D) image ${\boldsymbol F} \in {\mathbb R}^{n_x\times n_y}$; the image pixels are vectorized to $\boldsymbol{f}\in{\mathbb R}^{n}$, where $n = n_xn_y$.
Under a wavelet basis ${\boldsymbol T}\in\mathbb{R}^{n\times n}$,  $\boldsymbol{f}$ can be expressed as
${\boldsymbol{ f}}  = {\boldsymbol T} {\boldsymbol x}$, where the columns of $\boldsymbol{T}$ represent the orthonormal wavelet basis vectors, and $\xv$ represents the wavelet coefficients of $\fv$. We here focus the discussion on a wavelet transform, but the same type of tree-based model may be applied to a \emph{block} discrete cosine transform (DCT) representation \cite{Xiong96SPL,He10SPL}. In Section \ref{Sec:Result} experimental results are shown based on both a wavelet and block-DCT tree-based decomposition. However, for clarity of exposition, when presenting the model, we provide details in the context of the wavelet transform.

The measurement model for $\fv$ is assumed manifested in terms of $m$ linear projections, defined by the rows of ${\boldsymbol H} \in {\mathbb R}^{m\times n}$. Assuming measurement noise $\nv$, we have:

\begin{equation} \label{eq:yHf}
{\boldsymbol{y}}= {\boldsymbol{H f}} + \boldsymbol{n}={\boldsymbol{\Psi x}} + \boldsymbol{n}\end{equation}
where $\boldsymbol{\Psi}=\boldsymbol{H}{\boldsymbol T}$. In compressive sensing (CS) \cite{Candes05ITT} $m\ll n$, and it is desirable that the rows of $\boldsymbol{H}$ are incoherent with the columns of $\boldsymbol{T}$, such that $\boldsymbol{\Psi}$ is a dense matrix \cite{Candes05ITT,Donoho06ITT,Baraniuk07SPM}.
If one directly observes the image pixels, which we will consider in the context of image-denoising applications, ${\boldsymbol{H}}= {\boldsymbol I}$, with ${\boldsymbol I}$ symbolizing the identity matrix.

Assuming i.i.d. Gaussian noise with precision $\alpha_0$, we have
\begin{equation}\label{eq:GaussNoise}
{\boldsymbol{y}}|\xv \sim {\cal N}(\boldsymbol{\Psi x}, \alpha_0^{-1}{\boldsymbol I}).
\end{equation}
The assumption of i.i.d. Gaussian noise may be inappropriate in some settings, and in Section \ref{Sec:Spiky} we revisit the noise model.

\subsection{Modeling the wavelet scaling coefficients}
\label{Sec:model_tree}

In a wavelet decomposition of a 2D image, there are three sets of trees: one associated with high-high (HH) wavelet coefficients, one for high-low (HL) wavelet coefficients, and one for low-high (LH) coefficients \cite{Mallat}. In addition to the HH, HL and LH wavelet coefficients, there is a low-low (LL) scaling coefficient subimage, which does \emph{not} have an associated tree; the LL scaling coefficients constitute a low-frequency representation of the original image.

The scaling coefficients are in general {\em not} sparse (since the original image is typically not sparse in the native pixel basis), and hence the scaling coefficients are modeled as
\begin{equation}
x_{0,i}|\tau_0,\alpha_0 \sim {\cal N}(0,\tau_0^{-1}\alpha_0^{-1}), \label{eq:tree1}
\end{equation}
where $\tau_0$ controls the ``global'' scale of the scaling coefficients, and $\alpha_0$ is the noise precision as above ($\tau_0$ represents the \emph{relative} precision of the scaling coefficients, with respect to the noise precision $\alpha_0$). 
Non-informative gamma priors are placed on $\tau_0$ and $\alpha_0$, $i.e.$, $\mbox{Gamma}(10^{-6},10^{-6})$.

\subsection{Compressiblility of wavelet coefficients\label{sec:compressibility}}

Let $x_{\ell,i}$ represent coefficient $i$ at level $\ell$ in the wavelet tree (for either a HH, HL, or LH tree). To make connections to existing shrinkage priors and regularizers (\emph{e.g.}, adaptive Lasso \cite{ZouAdaLasso}), we consider a generalization of the double-exponential (Laplace) prior
\cite{Tibshirani94Lasso,Park08BLASSO}. In Section \ref{sec:Levy} we discuss that this model corresponds to one example of a broad class of shrinkage priors based on L\'{e}vy processes, to which the model may be generalized.

Coefficient
$x_{\ell,i}$ is assumed drawn from the distribution
\begin{eqnarray}
&&\frac{1}{2}\sqrt{\frac{\tau_\ell}{\gamma_{\ell,i}}} \exp(-|x_{\ell,i}|\sqrt{\frac{\tau_\ell}{\gamma_{\ell,i}}}) \nonumber\\
&=& \int{\cal N}(x_{\ell,i};0,\tau_\ell^{-1}\alpha^{-1}){\rm InvGa}(\alpha; 1,(2\gamma_{\ell,i})^{-1}) d \alpha  \label{eq:double}
\end{eqnarray}
Parameter $\tau_\ell >0$ is a ``global'' scaling for all wavelet coefficients at layer $\ell$, and $\gamma_{\ell,i}$ is a ``local'' weight for wavelet coefficient $i$ at that level. We place a gamma prior $\mbox{Gamma}(a,b)$ on the $\gamma_{\ell,i}$, and as discussed below one may set the hyperparameters $(a,b)$ to impose that most $\gamma_{\ell,i}$ are small, which encourages that most $x_{\ell,i}$ are small. Inference of a point estimate for the model parameters, via seeking to maximize the log posterior, one observes that the log of the prior in (\ref{eq:double}) corresponds to adaptive Lasso regularization \cite{ZouAdaLasso}.

The model (\ref{eq:double}) for wavelet coefficient $x_{\ell,i}$ may be represented in the hierarchical form
\begin{eqnarray}
x_{\ell,i}| \tau_{\ell},\alpha_{\ell,i}&\sim&\mathcal{N}(0,\tau_\ell^{-1}\alpha_{\ell,i}^{-1})\nonumber\\
 \alpha_{\ell,i}|\gamma_{\ell,i} &\sim& \mbox{InvGa}(1,(2\gamma_{\ell,i})^{-1})\label{eq:basic}\\
\gamma_{\ell,i}&\sim& \mbox{Ga}(a,b),\nonumber
\end{eqnarray}
introducing latent variables $\{\alpha_{\ell,i}\}$, rather than marginalizing them out as in (\ref{eq:double}); a 
vague/diffuse gamma prior is again placed on the scaling parameters $\tau_\ell$. While we have introduced many new latent variables $\{\alpha_{\ell,i}\}$, the form in (\ref{eq:basic}) is convenient for computation, and with appropriate choice of $(a,b)$ most $\{\alpha_{\ell,i}\}$ are encouraged to be large. The large $\alpha_{\ell,i}$ correspond to small $x_{\ell,i}$, and therefore this model imposes that most $x_{\ell,i}$ are small, $i.e.$, it imposes compressibility. Below we concentrate on the model for $\{\gamma_{\ell,i}\}$.

\subsection{Shrinkage priors versus spike-slab priors}

There are models of coefficients like $\xv$ that impose exact sparsity \cite{Spike-slab05,Schniter,He09SPT}, while the model in (\ref{eq:basic}) imposes compressibility (many coefficients that are small, but not exactly zero). Note that if a component of $\xv$ is small, but is set exactly to zero via a sparsity prior/regularizer, then the residual will be attributed to the additive noise $\nv$. This implies that the sparsity-promoting priors will tend to over-shrink, under-estimating components of $\xv$ and  over-estimating the noise variance. This partially explains why we have consistently found that compressibility-promoting priors like (\ref{eq:basic}) often are more effective in practice on real (natural) data than exact-sparsity-promoting models, like the spike-slab setup in \cite{He09SPT}. These comparisons are performed in detail in Section \ref{Sec:Result}. In Section \ref{sec:other_shrink} we make further connections shrinkage priors developed in the literature.

\section{Hierarchical Tree-Based Shrinkage\label{sec:TreeShrink}}

\subsection{Hierarchical gamma shrinkage}

\begin{figure}[ht!]
  \centering
  \includegraphics[scale = 1.0]{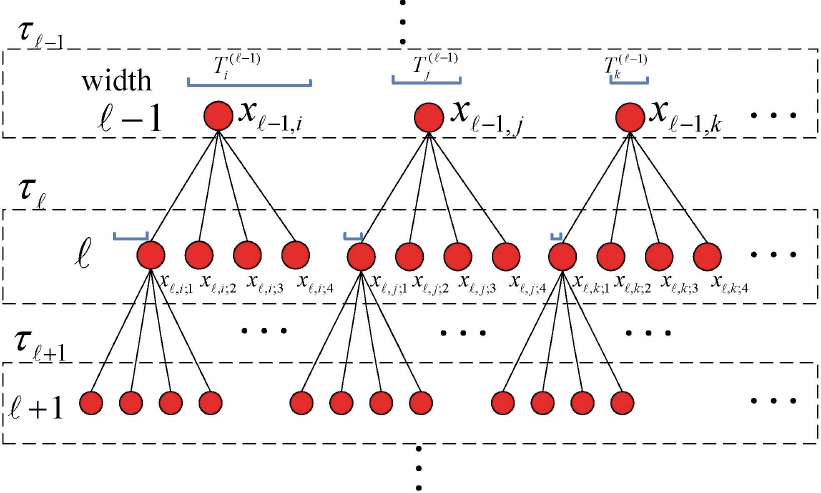}\\
\caption{\small{Depiction of the layers of the trees, with $n_c=4$ children considered. Note that each parent coefficient at layer $\ell-1$ has a relative width $T_i^{(\ell-1)}$, and all widths at a given layer sum to one. The width of the parent is partitioned into windows of width $T_i^{(\ell-1)}/n_c$ for the children.}}
  \label{Fig:tree}

\end{figure}

If we simply impose the prior $\mbox{Ga}(a,b)$ on $\{\gamma_{\ell,i}\}$, with $(a,b)$ set to encourage compressibility on $\{\gamma_{\ell,i}\}$, we do not exploit the known tree structure of wavelets and what that tree imposes on the characteristics of the multi-layer coefficients \cite{Mallat,Crouse98SPT}. A wavelet decomposition of a 2D image yields a quadtree hierarchy
($n_c=4$ children coefficients under each parent coefficient), as shown in Figure \ref{Fig:tree}. A key contribution of this paper is development of a general shrinkage-based decomposition that respects the tree structure.

Let $\boldsymbol{x}$ represent all the coefficients in a wavelet decomposition of a given image, where ${\boldsymbol{x}}_\ell$ represents all coefficients across {all} trees at level $\ell$ in the wavelet hierarchy. We here consider all the wavelet trees for one particular class of wavelet coefficients ($i.e.$, HH, HL or LH), and the model below is applied independently as a prior for each. Level $\ell=0$ is associated with the scaling coefficients, and $\ell=1, 2, \dots$ correspond to levels of wavelet coefficients, with $\ell=1$ the root-node level. We develop a prior that imposes shrinkage \emph{within} each layer $\ell$, while also imposing structure \emph{between} consecutive layers. Specifically, we impose the \emph{persistence} property \cite{Crouse98SPT} of a wavelet decomposition of natural imagery: if wavelet coefficient $i$ at level $\ell-1$, denoted $x_{\ell-1,i}$, is large/small, then its children coefficients $\{x_{\ell,i;j}\}_{j=1,n_c}$ are also likely to be large/small.

A parameter $\gamma_{\ell,i}$ is associated with each node $i$ at each level $\ell$ in the tree (see Figure \ref{Fig:tree}). Let $\{\gamma_{\ell-1,i}\}$ represent the \emph{set} of parameters for all coefficients at layer $\ell-1$. The parameters  $\{\gamma_{\ell,i;j}\}_{j=1,n_c}$ represent the $n_c$ \emph{children} parameters under \emph{parent} parameter $\gamma_{\ell-1,i}$.


If there are $n_1$ wavelet trees associated with a given 2D image, there are $n_1$ root nodes at the top layer $\ell=1$.
For each of the coefficients associated with the root nodes, corresponding shrinkage parameters are drawn
\begin{equation}
\gamma_{1,i}\sim \mbox{Ga}(1/n_1,b)
\end{equation}
which for large $n_1$ encourages that most $\gamma_{1,i}$ will be small, with a few outliers; this is a ``heavy-tailed'' distribution for appropriate selection of $b$, where here we set $b=1$ (the promotion of compressibility is made more explicit in Section \ref{sec:Levy}).

Assuming $n_{\ell-1}$ nodes at layer $\ell-1$, we define
an ``increment length'' $T_i^{(\ell-1)}$ for node $i$ at level $\ell-1$:
\begin{equation}\label{eq:norm}
T_i^{(\ell-1)}=\gamma_{\ell-1,i}/\sum_{i^\prime=1}^{n_{\ell-1}} \gamma_{\ell-1,i^\prime}
\end{equation}
To constitute each child parameter $\gamma_{\ell,i;j}$, for $j=1,\dots,n_c$, we independently draw
\begin{equation}
\gamma_{\ell,i;j}\sim \mbox{Ga}(T_i^{(\ell-1)}/n_c,1)
\end{equation}
Because of the properties of the gamma distribution, if $T_i^{(\ell-1)}$ is \emph{relatively} large ($i.e.$, $\gamma_{\ell-1,i}$ is relatively large), the children coefficients are also encouraged to be relatively large, since the integration window $T_i^{(\ell-1)}/n_c$ is relatively large; the converse is true if $T_i^{(\ell-1)}$ is small.

The use of the term ``increment'' is tied to the generalization of this model within the context of L\'{e}vy processes, discussed in Section \ref{sec:Levy}, where further justification is provided for associating increment lengths with shrinkage paramerers $\{\gamma_{\ell,i}\}$.

%
%
%
%

\subsection{Dirichlet simplification}

In (\ref{eq:basic}), the parameter $\tau_\ell$ provides a scaling of all precisions at layer $\ell$ of the wavelet tree. Of most interest to defining which wavelet coefficients have large values are the \emph{relative} weights of $\alpha_{\ell,i}$, with the relatively small elements corresponding to wavelet coefficients with large values. Therefore, for computational convenience, we consider scaled versions of $\gamma_{\ell,i}$, with the relatively large $\gamma_{\ell,i}$ associated with the relatively small-valued $\alpha_{\ell,i}$. Specifically, consider the normalized vector
\begin{equation}
\tilde{\boldsymbol{\gamma}}_{\ell,i} = \boldsymbol{\gamma}_{\ell,i}/\sum_{i^{\prime}}\gamma_{\ell,i^{\prime}},\end{equation}
which corresponds to a draw from a Dirichlet distribution:
\begin{equation} \tilde{\boldsymbol{\gamma}}_{\ell}\sim \mbox{Dir}(\beta_1^{(\ell)},\dots,\beta_{n_\ell}^{(\ell)}) \label{Eq:gamma_beta}
\end{equation}
where {$\beta_i^{(\ell)}=T_{pa(\ell,i)}^{(\ell-1)}/n_c$}, and $pa(\ell,i)$ is the parent of the $i$th coefficient at layer $\ell$.
The last equation in (\ref{eq:basic}) is replaced by (\ref{Eq:gamma_beta}), and all other aspects of the model remain unchanged; the ${\boldsymbol{\tilde{\gamma}}}_\ell$ are \emph{normalized} parameters, with the scaling of the parameters handled by $\tau_\ell$ at layer $\ell$.
One advantage of this normalization computationally is that all elements of $\tilde{\boldsymbol{\gamma}}_{\ell}$ are updated as a block, rather than one at a time, as implied by the last equation in (\ref{eq:basic}).

The model for parameters ${\boldsymbol{\tilde{\gamma}}}_\ell$ at each level of the tree may be summarized as follows. At the root $\ell=1$ layer
\begin{equation}
\boldsymbol{\tilde{\gamma}}_1\sim\mbox{Dir}(1/{n_1},\dots,1/{n_1})\end{equation}
If $\boldsymbol{\tilde{\gamma}}_{\ell-1}$ are parameters at level $\ell-1$, then the child parameters follow
(\ref{Eq:gamma_beta}), with {$\beta_i^{(\ell)}=T^{(\ell-1)}_{pa(\ell,i)}/n_c = \tilde{\gamma}_{\ell-1,pa(\ell,i)}$}.

\section{Multiscale L\'{e}vy Process Perspective\label{sec:Levy}}

\subsection{L\'{e}vy process and infinite divisibility\label{sec:infdiv}}

A L\'{e}vy process \cite{ApplebaumLevy} $G(t)$ is a continuous-time stochastic process
with $G(0)=0$, and with increments that are independent and stationary. Specifically, for any $t$, an increment of width $\Delta$ is defined by $G(t+\Delta)-G(t)$, and the distribution of $G(t+\Delta)-G(t)$ depends only on $\Delta$. Further, the random variables associated with non-overlapping increments are independent.
While this stochastic process can take values in a multidimensional space, we consider $G(t)$ on the real line.
The characteristic equation of the random variable $G(t=T)\in\mathbb{R}$ is
\begin{eqnarray}
&&\mathbb{E}(e^{i u G(T)})=\exp[T\{i\delta u-\frac{1}{2}u^2\Sigma \nonumber\\
&&\hspace{0.6cm}+\int_{\mathbb{R}-\{0\}}[e^{iuy}-1-iuy 1(|y|\leq 1)]\nu(dy)\}],
\end{eqnarray}
where $\nu$ is a L\'{e}vy measure satisfying
$\int_{\mathbb{R}-\{0\}}(|y|^2 \wedge 1)\nu(dy)<\infty
$. The terms $\delta\in\mathbb{R}$ and $\Sigma\in\mathbb{R}_+$ correspond to, respectively, the drift and variance {per unit time} of Brownian motion over time $[0,T]$,
and at time $T$, the Brownian contribution to $G(T)$ is drawn from a normal with mean $\delta T$, and variance $\Sigma T$: $G_B(T)\sim\mathcal{N}(\delta T, \Sigma T)$.
The term associated with L\'{e}vy measure $\nu$ corresponds to a Poisson process (PP) \cite{Kingman02} over time period $[0,T]$, with PP rate
$\nu(dy)$; the PP contribution to $G(T)$, denoted $G_{PP}(T)$, is the integral of the PP draw over $[0,T]$. The latter corresponds to a 
pure-jump processes, and $G(T)=G_B(T)+G_{PP}(T)$.
Importantly for our purposes, the parameter $T$ controls the mean and variance on $G(T)$, with these terms increasing with $T$.

The infinitely divisible nature of a L\'{e}vy process \cite{ApplebaumLevy} allows us to express $G(T)=\sum_{i=1}^n G_i$ for any $n$, where each $G_i$ is an \emph{independent} draw from the same L\'{e}vy process as above, but now with time increment $T/n$. Each of the $G_i$ is the sum of a Brownian motion term, with drift $\delta T/n$ and variance $\Sigma T/n$, and a jump term with rate $T\nu(dy)/n$. As first discussed in \cite{Polson10shrinkglobally} by choosing particular settings of $\delta$, $\Sigma$ and $\nu$, the vector $\{G_i\}$ may be used to impose sparsity or compressibility.

Specifically, assume that we wish to impose that the components of $\boldsymbol{x}\in\mathbb{R}^n$ are compressible. Then the $i$th component $x_i$ may be assumed drawn $x_i\sim\mathcal{N}(0,\tau G_i)$, where $\tau$ is a ``global'' scaling, independent of $i$. Since only a small subset of the $\{G_i\}_{i=1,n}$ are \emph{relatively} large for an appropriate L\'{e}vy process (with $\delta=\Sigma=0$, and a L\'{e}vy measure $\nu$ characterized by a small number of jumps \cite{Polson10shrinkglobally}), most components $x_i$ are encouraged to be small. This global scaling $\tau$ and local $G_i$ looks very much like the class of models in (\ref{eq:basic}), which we now make more precise.

\subsection{Tree-based L\'{e}vy process}

We develop a hierarchical L\'{e}vy process, linked to the properties of tree-based wavelet coefficients; from that, we will subsequently make a linkage to the hierarchical prior developed in Section \ref{sec:TreeShrink}. At each layer $\ell$ in the tree, we impose a total time window $T=1$. An underlying continuous-time L\'{e}vy process $G_\ell (t)$ is assumed for each layer $\ell$, for $t\in[0,T]$. Assuming $n_\ell$ wavelet coefficients at layer $\ell$, we consider $n_\ell$ positive increment lengths $\{T_i^{(\ell-1)}\}_{i=1,n_\ell}$, with $\sum_{i=1}^{n_\ell} T_i^{(\ell-1)}=1$. The $i$th increment of the L\'{e}vy process at layer $\ell$, $i\in\{1,\dots,n_\ell\}$, is defined as $G_{\ell,i}=G_\ell (\sum_{j=1}^{i} T_j)-G_\ell (\sum_{j=1}^{i-1} T_j)$.

The L\'{e}vy-process-based tree structure, analogous to that considered in Section \ref{sec:TreeShrink}, is defined as follows. At the root layer $\ell=1$, with $n_1$ wavelet coefficients, we define $T_i^{(1)}=1/n_1$ for all $i\in\{1,\dots,n_1\}$. From L\'{e}vy process $G_1(t)$ we define increments $G_{1,i}$, $i=1,\dots, n_1$.

Assume increments $G_{\ell-1,i}$, $i=1,\dots, n_{\ell-1}$ at layer $\ell-1$. The lengths of the increments at layer $\ell$ are defined analogous to (\ref{eq:norm}), specifically
\begin{equation}
T_i^{(\ell-1)}=|G_{\ell,i}|/\sum_{i^\prime}^{n_\ell-1} |G_{\ell,i^\prime}|
\end{equation}
The $n_c$ children of coefficient $i$ at layer $\ell-1$ have L\'{e}vy increment widths $T_i^{(\ell-1)}/n_c$.
Therefore, if an increment $G_{\ell,i}$ is relatively large/small compared to the other coefficients at layer $\ell$, then the window lengths of its children coefficients, $T_i^{(\ell-1)}/n_c$, are also relatively large/small. Since the expected relative strength of a L\'{e}vy increment is tied to the width of the associated increment support, this construction has the property of imposing a persistence in the strength of the wavelet coefficients across levels $\ell$ (relatively large/small parent coefficients enourage relative large/small children coefficients).

\subsection{Shrinkage and sparsity imposed by different L\'{e}vy families\label{sec:LevyFamilies}}

Each choice of L\'{e}vy parameters $(\delta,\Sigma,\nu)$ yields a particular  hierarchy of increments $\{G_{\ell,i}\}$. In the context of imposing compressibility or sparsity, we typically are interested in $\delta=\Sigma=0$, and different choices of $\nu$ yield particular forms of sparsity or shrinkage. For example, consider the gamma process, with
\begin{equation}\label{eq:gamma}
\nu(dy)=y^{-1}\exp(-yb) dy
\end{equation}
defined for $y>0$. Using notation from above, for a L\'{e}vy process $G_\ell (t)$ drawn from a gamma process, the increment is a random variable $G_{\ell,i}\sim \mbox{Ga}(T_i^{(\ell-1)},b)$. Therefore, the model considered in Section \ref{sec:TreeShrink} is a special case of the hierarchical tree-based L\'{e}vy process, with a gamma process L\'{e}vy measure. The increments $G_{\ell,i}$ correspond to the $\gamma_{\ell,i}$ in (\ref{eq:basic}).

Note the singularity in (\ref{eq:gamma}) at $y=0$, and also the heavy tails for large $y$. Hence, the gamma process imposes that most $G_{\ell,i}$ will be very small, since the gamma-process is concentrated around $y=0$, but the heavy tails will yield some large $G_{\ell,i}$; this is why the increments of appropriate L\'{e}vy processes have close connections to compressibility-inducing priors \cite{Polso12Levy}.

There are many choices of L\'{e}vy measures one may consider; the reader is referred to \cite{Polso12Levy,Polson10shrinkglobally,Wolp:Clyd:Tu:2011} for a discussion of these different L\'{e}vy measures and how they impose different forms of sparsity/compressibility. We have found that the gamma-process construction for $\{\alpha_{\ell,i}\}$, which corresponds to Section \ref{sec:TreeShrink}, works well in practice for a wide range of problems, and therefore it is the principal focus of this paper.

We briefly note another class of L\'{e}vy measures that is simple, and has close connections to related models, discussed in Section \ref{sec:related}. Specifically, consider $\nu(dy)=\nu^+H(dy)$, where $\nu^+$ is a positive constant and $H(dy)$ is a probability density function, $i.e.$, $\int H(dy)=1$. This corresponds to a compound Poisson process, for which $J\sim \mbox{Pois}(\nu^+)$ atoms are situated uniformly at random within the window $[0,T]$, and the amplitude of each atom is drawn i.i.d. from $H$. In this case $G_\ell(t)$ is a jump process that can be positive and negative, if $H$ admits random variables on the real line ($e.g.$, if $H$ corresponds to a Gaussian distribution). Rather than using this L\'{e}vy process to model $\gamma_{\ell,i}$, which must be positive, the increments of $G_\ell(t)$ are now used to directly model the wavelet coefficients $x_{\ell,i}$. We demonstrate in Section \ref{sec:OtherTree} that this class of hierarchical L\'{e}vy constructions has close connections to existing models in the literature for modeling wavelet coefficients.

\section{Connections to Previous Work\label{sec:related}}

\subsection{Other forms of shrinkage\label{sec:other_shrink}}

Continuous shrinkage priors, like (\ref{eq:basic}) and its hierarchical extensions developed here, enjoy advantages over conventional discrete mixture \cite{mitchell1988bayesian,george1993variable} ($e.g.$, spike-slab) priors.  First, shrinkage may be more appropriate than selection in many scenarios (selection is manifested when exact sparsity is imposed). For example, wavelet coefficients of natural images are in general $p\mbox{-}$compressible \cite{cevher2009learning}, with many small but non-zero coefficients.  Second, continuous shrinkage avoids several computational issues that may arise in the discrete mixture priors (which impose explicit sparsity) \cite{carvalho2009handling}, such as a combinatorial search over a large model space, and high-dimensional marginal likelihood calculation. Shrinkage priors
often can be characterized as Gaussian scale mixtures (GSM), which naturally lead to simple block-wise Gibbs sampling updates with better mixing and convergence rates in MCMC \cite{Polson10shrinkglobally,Armagan12}.  Third, continuous shrinkage reveals close connections between Bayesian methodologies and frequentist regularization procedures. For example, in Section \ref{sec:compressibility} we noted the connection of the proposed continuous-shrinkage model to 
adaptive Lasso. Additionally, the iteratively reweighted $\ell_2$ \cite{daubechies2010iteratively} and $\ell_1$ \cite{candes2008enhancing} minimization schemes could also be derived from the EM updating rules in GSM and Laplace scale mixture (LSM) models  \cite{ garrigues2010group,zhang2012ep}, respectively.

Recently, aiming to better mimic the marginal behavior of discrete mixture priors, \cite{Polson10shrinkglobally, Polso12Levy} present a new global-local (GL) family of GSM priors,
\begin{eqnarray}\label{eq: GL}
x_{i}|\tau,\lambda_{i}\sim \mathcal{N}(0, \tau\lambda_{i}),\quad \lambda_{i}\sim f, \quad \tau \sim g,
\end{eqnarray}
where $f$ and $g$ are the prior distributions of $\lambda_i$ and $\tau$; as noted above, such priors have been an inspiration for our model.
The local shrinkage parameter $\lambda_{i}$ and global shrinkage parameter $\tau$ are able to offer sufficient flexibility in high-dimensional settings. There exist many options for the priors on the Gaussian scales $\lambda_{i}$,  for example  in the three parameter beta normal $\mathcal{TPBN}(a,b,\tau)$ case \cite{Armagan12},
\begin{equation}
x_{i}\sim \mathcal{N}(0, \tau\lambda_{i}), \lambda_{i}\sim \mathrm{Ga}(a, \gamma_{i}), \gamma_{i}\sim \mathrm{Ga}(b,1)
\end{equation}
The horseshoe prior \cite{Carvalho_horseshoe} is a special case of ${\cal TPBN}$ with $a=1/2$, $b=1/2$, $\tau=1$.

GSM further extends to the LSM \cite{garrigues2010group} by adding one hierarchy in \eqref{eq: GL}: $\lambda_{i}\sim \mathrm{Ga}(1,\gamma_{i})$, $\gamma_{i}\sim f$, which captures higher order dependences.  As a special case of LSM, the normal exponential gamma (NEG) prior \cite{griffin2011bayesian} provides a Bayesian analog of adaptive lasso \cite{ZouAdaLasso}. The Dirichlet Laplace (DL) prior employed here can be interpreted as a non-factorial LSM prior which resembles point mass mixture prior in a joint sense, with frequentist optimality properties well studied in \cite{Bhattacharya12}.

None of the above shrinkage priors take into account the multiscale nature of many compressible coefficients, like those associated with wavelets and the block DCT. This is a key contribution of this paper.

\subsection{Other forms of imposing tree structure\label{sec:OtherTree}}

There has been recent work exploiting the wavelet tree structure within the context of compressive sensing (CS) \cite{He09SPT,He10SPL,Som12TSP}. In these models a two-state Markov tree is constituted, with one state corresponding to large-amplitude wavelet coefficients and the other to small coefficients. If a parent node is in a large-amplitude state, the Markov process is designed to encourage the children coefficients to also be in this state; a similar state-transition property is imposed for the small-amplitude state.

The tree-based model in \cite{He09SPT} is connected to the construction in Section \ref{sec:Levy}, with a compound Poisson L\'{e}vy measure. Specifically, consider the case for which $\delta=\Sigma=0$, and $\nu$ is a compound Poisson process with $\nu=\nu_\ell^+ \mathcal{N}(0,\zeta_\ell^{-1}\boldsymbol{I})$, where $\nu_\ell^+$ a positive real number and $\zeta_\ell$ a precision that depends on the wavelet layer $\ell$. Priors may be placed on $\nu_\ell^+$ and $\zeta_\ell$. At layer $\ell$, $J_\ell\sim\mbox{Pois}(\nu_\ell^+)$ ``jumps'' are drawn, and for each a random variable is drawn from $\mathcal{N}(0,\zeta_\ell^{-1}\boldsymbol{I})$; the jumps are placed uniformly at random between [0,1], defining $G_\ell(t)$. Along the lines discussed in Section \ref{sec:Levy}, at the root layer of the L\'{e}vy tree, each of the $n_1$ wavelet coefficients is assigned a distinct increment of length $1/n_1$. As discussed at the end of Section \ref{sec:LevyFamilies}, here the increments $G_{\ell,i}$ are used to model the wavelet coefficients directly. If one or more of the $J_1$ aforementioned jumps falls within the window of coefficient $i$ at layer $\ell=1$, then the coefficient is non-zero, and is a Gaussian distributed; if no jump falls within width $i$, then the associated coefficient is exactly zero. If a coefficient is zero, then all of its decendent wavelet coefficients are zero, using the construction in Section \ref{sec:Levy}. This may be viewed as a two-state model for the wavelet coefficients: each coefficient is either exactly zero, or has a value drawn from a Gaussian distribution. This is precisely the type of two-state model developed in \cite{He09SPT}, but viewed here from the perspective of a tree-based L\'{e}vy process, and particularly a compound Poisson process. In \cite{He09SPT} a compound Poisson model was not used, and rather a Markov setup was employed, but the basic concept is the same as above.

In \cite{Som12TSP} the authors developed a method similar to \cite{He09SPT}, but in \cite{Som12TSP} the negligible wavelet coeffecients were not shrunk exactly to zero. Such a model can also be viewed from the perspective of the L\'{e}vy tree, $\nu$ again corresponding to a compound Poisson process. However, now $\delta=0$ and $\Sigma\neq 0$. For an increment for which there are no jumps, the wavelet coefficient is only characterized by the Brownian motion term, and $\Sigma$ is associated with small but non-zero wavelet coefficients.

This underscores that the proposed tree-based shrinkage prior, which employs the gamma process, may be placed within the context of the L\'{e}vy framework of Section \ref{sec:Levy}, as can \cite{He09SPT,Som12TSP}, which use state-based models. The placement of all of these models in the L\'{e}vy framework provides a unification, and also suggests other types of priors that respect the tree-based structure, via different choices of $(\delta,\Sigma,\nu)$. This provides a significant generalization of \cite{Polson10shrinkglobally, Polso12Levy}, which first demonstrated that many shrinkage and sparsity priors may be placed within the context of the increments of a L\'{e}vy process, but did not account for the tree-based structure of many representations, such as wavelets and the block DCT.

\subsection{Connection to dictionary learning\label{sec:dict}}

In Section \ref{Sec:Result} we present experimental results on denoising images, and we make comparisons to an advanced dictionary-learning approach \cite{Zhou11AISTATS}. It is of interest to note the close connection of the proposed wavelet representation and a dictionary-learning approach. Within the wavelet decomposition, with $n_1$ root nodes, we have $n_1$ trees, and therefore the overall construction may be viewed as a ``forest." Each tree has $L$ layers, for an $L$-layer wavelet decomposition, and each tree characterizes a subregion of the image/data. Therefore, our proposed model can be viewed as dictionary learning, in which the dictionary building blocks are wavelets, and the tree-based shrinkage imposes form on the weights of the dictionary elements within one tree (corresponding to an image patch or subregion). Within the proposed gamma process construction, the ``local'' $\{\gamma_{\ell,i}\}$ account for statistical relationships between coefficients at layer $\ell$, the parent-child encoding accounts for relationships between consecutive layers, and the ``global'' $\tau_\ell$ also accounts for cross-tree statistical relationships.

\section{Generalized Noise Model\label{Sec:Spiky}}

In (\ref{eq:GaussNoise}) the additive noise was assumed to be i.i.d. zero-mean Gaussian with precision $\alpha_0$. In many realistic scenarios, the noise may be represented more accurately as a sum of a Gaussian and spiky term, and in this case we modify the model as
\begin{equation}
{\boldsymbol y}
= {\boldsymbol{\Psi}} {\boldsymbol x} + {\boldsymbol w} + {\boldsymbol n}, \label{Eq:Ytheta}
\end{equation}
where the spiky noise ${\boldsymbol w}$ is also modeled by a shrinkage prior:
\begin{eqnarray}
w_i|\nu,\zeta_i,\alpha_0 &\sim&  {\cal N}(0, \nu^{-1}\zeta_i^{-1}\alpha_0^{-1}), \nonumber\\
\zeta_i|p_i &\sim&  {\rm InvGa}(1, (2p_i)^{-1}), \nonumber \\
{\boldsymbol p} &\sim& {\rm Dir}(1/n,\dots, 1/n),\nonumber \\
\nu &\sim&  {\rm Ga}(e_0, f_0).
\end{eqnarray}
The Gaussian term is still modeled as ${\boldsymbol n}|\alpha_0 \sim  {\cal N}(0, \alpha_0^{-1} {\boldsymbol I})$.

In this case the noise is a L\'{e}vy process, with increments defined by the size of the pixels: the term $\boldsymbol{n}$ corresponds to the Brownian term, with $\boldsymbol{w}$ manifested by the jump term (with L\'{e}vy measure $\nu$). Interestingly, the L\'{e}vy process is now playing two roles: ($i$) to model in a novel hierarchical manner the tree structure in the coefficients $\boldsymbol{x}$, and ($ii$) to separately model the Gaussian plus spiky noise $\boldsymbol{w}+\boldsymbol{n}$; the particular L\'{e}vy triplet $(\delta,\Sigma,\nu)$ need not be the same for these two components of the model.

\section{Inference\label{sec:inference}}
\label{Sec:inference}

We developed an MCMC algorithm for posterior inference, as well as a deterministic variational algorithm to approximate the posterior.
The latter is easily adapted to develop an Expectation-Maximization (EM) algorithm to obtain MAP point estimates of the wavelet coefficients, and
recalls frequentist algorithms like adaptive Lasso \cite{ZouAdaLasso} and re-weighted $\ell_1$ \cite{candes2008enhancing}. Details on the inference update equations are provided in the Appendix, and below we summarize unique aspects of the inference associated with the particular model considered.

\subsection{MCMC inference}

The MCMC algorithm involves a sequence of Gibbs updates, where each latent variable is resampled conditioned on the rest. Most conditional updates are
conjugate, the important exception being the coefficients $\gamma_{\ell,i}$ at each level; these have conditionals
\begin{eqnarray}
p(\tilde{\gamma}_{\ell,i}|-)&\propto& {\rm InvGa}(\alpha_{\ell,i}|1,(2\tilde{\gamma}_{\ell,i}^{-1})){\rm Dir}({\boldsymbol{\tilde{\gamma}}}_{\ell}|{\boldsymbol{\tilde{\gamma}}}_{\ell-1}) \nonumber\\
&&\times{\rm Dir}({\boldsymbol{\tilde{\gamma}}}_{\ell+1}|\boldsymbol{\tilde{\gamma}}_{\ell}), \\
p({\gamma}_{\ell,i}|-)&\propto& {\rm GIG}\left(2, \frac{\sum_{j\neq i}{\gamma}_{\ell,j}}{\alpha_{\ell,i}}, \frac{{{\tilde{\gamma}}}_{\ell-1,pa(\ell,i)}}{n_c}-1\right)\nonumber\\
&&\times{\rm Dir}\left({\boldsymbol{\tilde{\gamma}}}_{\ell+1}|{\boldsymbol{\tilde{\gamma}}}_{\ell}\right) ~\sum_j{\gamma}_{\ell,j}.  \label{eq:Phi_i}
\end{eqnarray}
Recall that $\tilde\gamma_{\ell,i}$ is just $\gamma_{\ell,i}$ normalized.
We performed this update using a Metropolis step \cite{Hastings70MH}, proposing from the generalized-inverse-Gaussian (GIG) distribution specified by the first term, with the last two
terms determining the acceptance probability.
In our experiments, we observed acceptance rates of about $80\%$.

\subsection{Heuristic variational posterior approximation}

Here we deterministically approximate the posterior distribution $p(\cdot|{\boldsymbol{y}})$ by a 
parametric distribution $q(\cdot)$, which we then optimize
to match the true posterior distribution. We use a mean-field factorized approximation, assuming a complete factorization across the latent variables,
$q(\boldsymbol{\Theta}) = \prod_{i} q_i(\Theta_i)$.
We approximate the distribution of each wavelet coefficient $x_{\ell,i}$ as a Gaussian distribution with mean $\mu_{x_{\ell,i}}$ and variance
$\sigma^2_{\ell,i}$. The marginal distributions over $\alpha$ and $\tau_{\ell}$ were set to exponential distributions.
The $\boldsymbol{\tilde{\gamma}}_{\ell}$ vectors were set as Dirichlet distributions with parameter vector $(b_1,\cdots, b_{n_c})$.
To optimize these parameters, we used a heuristic very closely related to expectation propagation \cite{Min2001a}.
For any variable (say $x$),
the corresponding Gibbs update rule from the previous section gives its distribution conditioned on its Markov blanket. We look at this conditional distribution,
plugging in the the average configuration of the Markov blanket (as specified by the current settings of the
posterior approximations $q(\cdot)$).
We then calculate the relevant moments of resulting conditional distribution over $x$, and update the variational distribution
$q(x)$ to match these moments. We repeat this procedure, cycling over all variables, and seeking a compatible configuration of all parameters
(corresponding to a fixed point of our algorithm).
As mentioned earlier, the resulting update equations
recall various frequentist algorithms, and we list them in the
Appendix. Two important ones are:
\begin{eqnarray}
\langle x_{{\ell},i}\rangle &=&\langle\alpha_0\rangle \sigma^2_{x_{\ell,i}} {\boldsymbol \Psi}_k^T  \left({\boldsymbol y} - \sum^{n}_{l=1,l\neq k}{\boldsymbol \Psi}_l \langle x_{l}\rangle\right),\\
\sigma^{2}_{{\ell,i}} &=& \langle\alpha_0\rangle^{-1}(\langle\tau_{\ell}\rangle \langle\alpha_{\ell,i}\rangle + {\boldsymbol \Psi}_k^T {\boldsymbol \Psi}_k)^{-1},
\end{eqnarray}
where $k$ is the index of ${\boldsymbol{x}}$ corresponding to $x_{\ell,i}$, ${\boldsymbol \Psi}_k$ is the $k$th column of ${\boldsymbol \Psi}$, 
and $\langle \cdot\rangle$ denotes the expectation value of the entry in $\langle \hspace{0.5mm} \rangle$.
It is worth noting that updating the $\gamma$'s requires calculating the mean of the distribution specified in equation \eqref{eq:Phi_i} for the current
averages of the posterior approximation. Unfortunately, this has no closed form. One approach is to calculate an importance sampling-based
 Monte Carlo estimate of this quantity,
using the proposal distribution specified in the previous section. If $s$ samples were used to produce this estimate, we call the resulting algorithm
VB($s$). The high acceptance rate of samples from the GIG suggests that this is a reasonably accurate approximation to the intractable
distribution. This suggests a simpler approach where all terms are ignored except for the GIG, whose mean is used to update the $\gamma$'s.
In this case, the update rules become:
{\begin{equation}
\langle{\gamma}_{\ell,i}\rangle =  \frac{\sqrt{\langle\frac{1}{\alpha_{\ell,i}}\rangle \sum_{j\neq i}\langle\tilde{\gamma}_{\ell,j} \rangle} K_{{{\tilde{\gamma}}}_{\ell-1,pa(\ell,i)}/n_c}(\chi)}{\sqrt{2}K_{({{\tilde{\gamma}}}_{\ell-1,pa(\ell,i)}/n_c-1)}(\chi)}, \nonumber
\end{equation}}
where $K_p(\chi)$ is the modified Bessel function of the second kind, and $\chi=\sqrt{2\langle \frac{1}{\alpha_{\ell,i}}\rangle\sum_{j\neq i}\langle\tilde{\gamma}_{\ell,j} \rangle}$.
We call this approximation a-VB.
This approximation can also be interpreted from the perspective of a wavelet hidden Markov (HM) tree \cite{He09SPT}, in which the dependence is always from the parent-node, rather than the child-nodes.
Thus, the underlying dependence inside the HM tree is {\em single} direction, from parent to children.

\subsection{MAP estimates via Expectation-Maximization}

Closely related to the previous approach is an algorithm that returns MAP estimates of the wavelet coefficients, while (approximately) averaging out
all other variables. This maximization over $\gamma_{\ell,i}$ must also be done approximately, but is a straightforward modification of the variational
update of $\gamma_{\ell,i}$. From the properties of the GIG, the M-step becomes
\begin{equation}
{\gamma}_{\ell,i}= \frac{\eta+\sqrt{\eta^2+\frac{2\sum_{j\neq i}{\gamma}_{\ell,j}}{\alpha_{\ell,i}}}}{2},
\end{equation}
where $\eta =\frac{{{\tilde{\gamma}}}_{\ell-1,pa(\ell,i)}}{n_c}-2$.
The approximate {E-Step} remains unaffected.

\section{Numerical Results\label{Sec:Result}}

Our code is implemented in non-optimized MATLAB, with all results
generated on a laptop with a 2.7 GHz CPU and 8 GB RAM.
Parameters $\alpha_0$, $\tau_{\ell}$ and $\nu$ are all drawn form a broad gamma prior, $\mbox{Gamma}(10^{-6},10^{-6})$. When performing inference, all parameters are initialized at random. No parameter tuning has been performed.
We run VB and EM for $100$ iterations, while our MCMC inferences were based on runs of $5000$ samples with a discarded burn-in of $1000$ samples. These number of iterations were not optimized, and typically excellent results are obtained with far fewer.

\subsection{Details on CS implementation and comparisons}

We apply the proposed approach, which we refer to as a shrinkage hierarchical model (denoted  ``s-HM" in the following figures) to compressive sensing (CS).
In these experiments the elements of the projection matrix $\boldsymbol{H}\in\mathbb{R}^{m\times n}$ are drawn i.i.d. from $\mathcal{N}(0,1)$, and the ``CS ratio,'' denoted CSR, is $m/n$.

In addition to considering s-HM, we consider a ``flat'' version of the shrinkage prior, ignoring the tree structure (analogous to models in \cite{Polso12Levy}); for this case the prior on the wavelet coefficients is as in (\ref{eq:basic}). Additionally, we make comparisons to the following algorithms (in all cases using publicly available code, with default settings):

\begin{itemize}
\item TSW-CS \cite{He09SPT,He10SPL}, using code from \url{http://people.ee.duke.edu/~lcarin/BCS.html}
\item OMP \cite{Tropp07ITT}, using code from  \url{http://www.cs.technion.ac.il/~elad/software/}
\item $\ell_1$ magic with TV norm ($\ell_1$-TV) \cite{Candes05ITT}, using code from \url{http://users.ece.gatech.edu/~justin/l1magic/}
\item Bayesian compressive sensing (BCS) \cite{Ji08SPT}, using code from \url{http://people.ee.duke.edu/~lcarin/BCS.html}
\item linearized Bregman \cite{Yin08bregman}, using code from \url{http://www.caam.rice.edu/~optimization/linearized_bregman/}
\end{itemize}

\subsection{CS and wavelet-based representation}

We consider images of size $128\times 128$, and hence $n=16,384$.
We use the ``Daubechies-4" wavelets \cite{Mallat}, with scaling coefficients (LL channel) of size $8\times 8$. In all Bayesian models, the posterior mean results are used for the reconstructed image.

\begin{figure}[ht!]
    \centering
    \includegraphics[scale = 0.25]{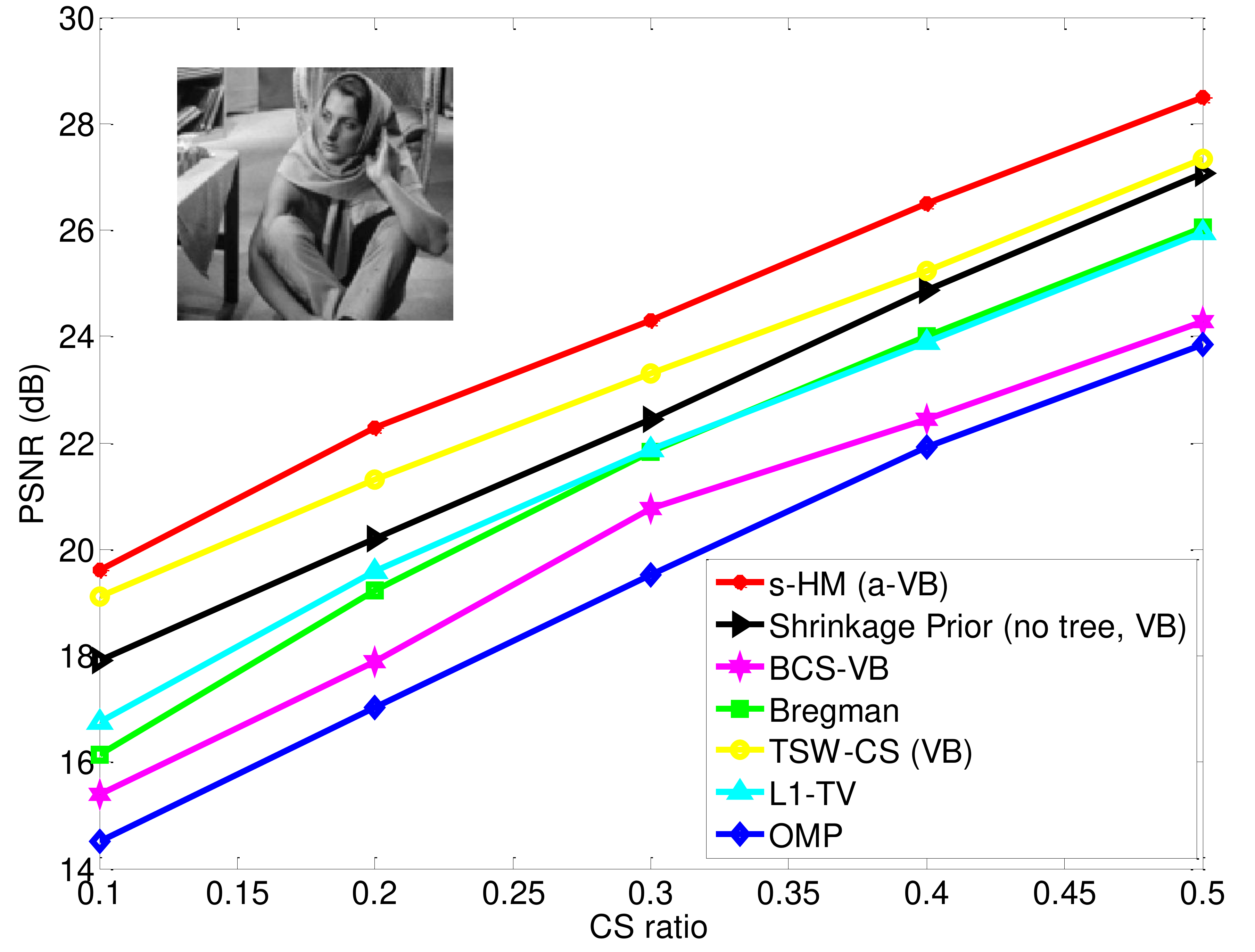}\\
    \caption{\small{PSNR comparison of various algorithms for compressive sensing with wavelet-based image representation.}} 
    \label{Fig:BarbaraPSNR}
\end{figure}

Figure \ref{Fig:BarbaraPSNR} plots the reconstruction peak signal-to-noise-ratio (PSNR) versus CS ratio for ``Barbara".
We see that
a) the proposed s-HM tree model provides best results, and
b) models (s-HM and TSW-CS) with tree structure are better than those without tree structure, consistent with the theory in \cite{BaraniukMCS}.

Similar observations hold for other widely considered images, viz.\ ``Lena" and ``Cameraman" (omitted for brevity). The PSNR of the reconstructed images by the proposed s-HM tree model is about 1dB higher than TSW-CS.
It is worth noting that even {\em without} the tree structure, the proposed local-global shrinkage prior performs well.
The other tree-based model, TSW-CS, has a prior that, via a spike-slab construction, encourages explicit sparsity; this seems to undermine results relative to the proposed shrinkage-based tree model, which we examine next.

\subsubsection{Estimated Wavelet Coefficients}
\label{Sec:mixing}

\begin{figure}[ht!]
  \centering
  \includegraphics[height = 4cm, width = 9cm]{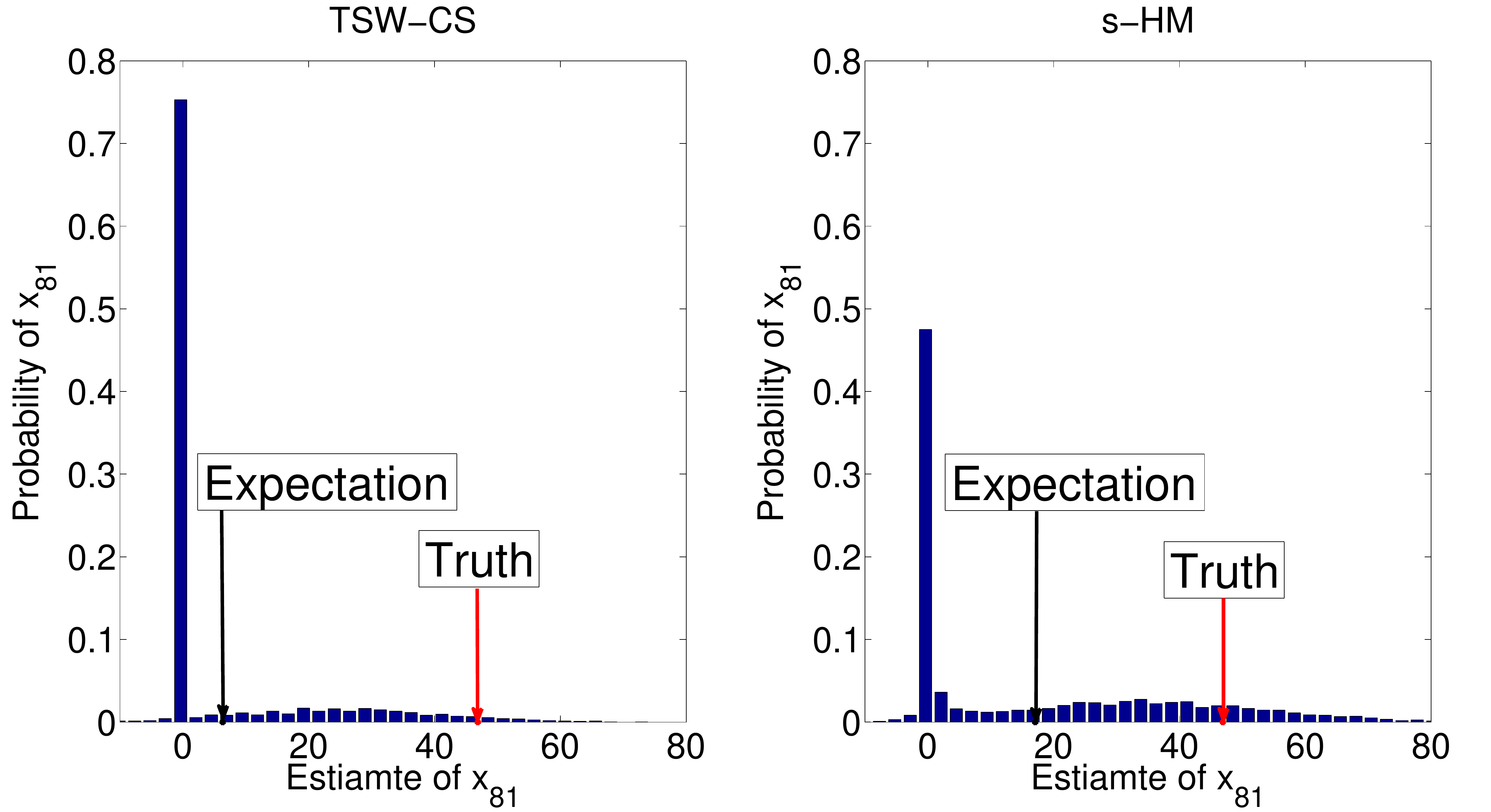}\\
  \caption{\small{Distribution of estimated $x_{81}$ by proposed s-HM model (right) and TSW-CS (left).}}
  \label{Fig:Mix81}
\end{figure}

We compare the posterior distribution (estimated by MCMC) of the estimated wavelet coefficients from the proposed model versus TSW-CS \cite{He09SPT}, in which the tree-based Markovian spike-slab model is used.
We consider ``Barbara" with a CSR=$0.4$.
An example of the posterior distribution for a typical wavelet coefficient is shown in Figure \ref{Fig:Mix81}, comparing the proposed s-HM with TSW-CS. Note that s-HM puts much more posterior mass around the true answer than TSW-CS.

Because of the spike-slab construction in TSW-CS, each draw from the prior has explicit sparsity, while each draw from the prior for the shrinkage-based s-HM always has all coefficients being non-zero. While TSW-CS and s-HM both impose and exploit the wavelet tree structure, our experiments demonstrate the advantage of s-HM because it imposes
compressibility rather than explicit sparsity.
Specifically, note from Figure \ref{Fig:Mix81} that the coefficient inferred via TSW-CS is shrunk toward zero more than the s-HM, and TSW-CS has little posterior mass around the true coefficient value. This phenomenon has been observed in numerous comparisons. The over-shrinkage of TSW-CS implies an increased estimate of the noise level $\boldsymbol{n}$ for that algorithm (the small wavelet coefficients are shrunk to exactly zero, and the manifested residual is incorrectly attributed to $\nv$).
This is attributed as the key reason the s-HM algorithm consistently performs better than TSW-CS for CS inversion.

\subsubsection{Robustness to Noise}

\begin{figure}[ht!]

        \centering
         \includegraphics[scale = 0.4]{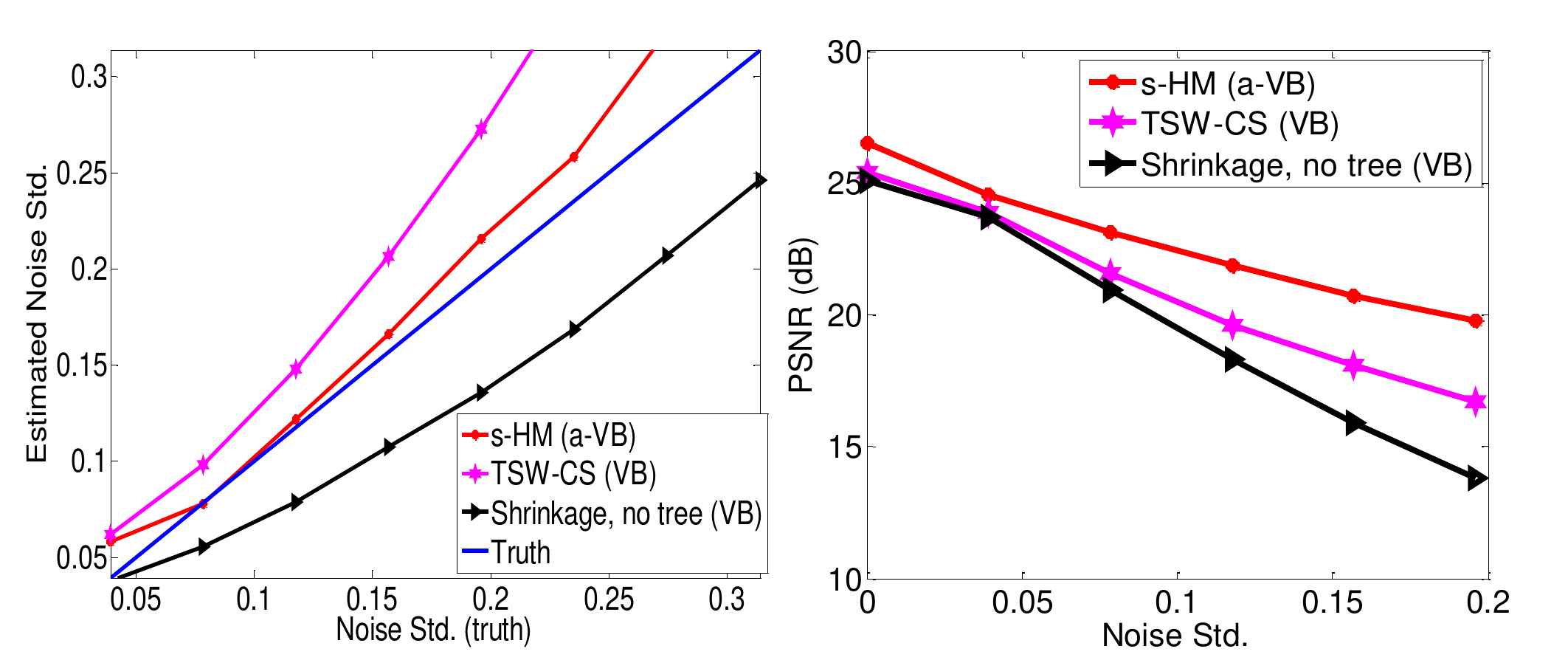}\\
         \caption{\small{Left: inferred noise standard deviation plotted versus truth. Right: reconstructed PSNR of ``Barbara" with different algorithms, at CS ratio = 0.4 under various noise levels. }}
          \label{Fig:Noise}

 \end{figure}

Based upon the above discussions, it is expected that s-HM will more accurately estimate the variance of additive noise $\boldsymbol{n}$, with TSW-CS over-estimating the noise variance.
We study the robustness of CS reconstruction under different levels of measurement noise, by adding zero-mean measurement Gaussian noise when considering ``Barbara''; the standard deviation here takes values in
[0, 0.31] (the pixel values of the original image were normalized to lie in [0,1]). Figure \ref{Fig:Noise} (right) plots the PSNR of the reconstruction results of the three best
algorithms from Figure \ref{Fig:BarbaraPSNR} (our tree and flat shrinkage models, and TSW-CS).
We see that the proposed s-HM tree model is most robust to noise, with its difference from TSW-CS growing with noise level.

Figure \ref{Fig:Noise} (left) plots the noise standard derivation inferred by the proposed models and TSW-CS (mean results shown). We see the tree structured s-HM model provides the best estimates of the noise variance, while
the TSW-CS overestimates it. The noise-variance estimated by the flat shrinkage model is under-estimated (apparently the absence of the wavelet tree structure in the flat model causes the model to attribute some of the noise incorrectly to wavelet coefficients).

\begin{table*}[ht!]
\caption{\small{Reconstruction PSNR (dB) with different inferences of ``Barbara" for the proposed s-HM model.}}
\centering
\footnotesize
\begin{tabular}{|c|c|c|c|c|c|c|c|}
\hline CS ratio 
& MCMC  & VB(500)& VB(100) & VB(50)& a-VB  & EM & VB(TSW-CS) \\
\hline 0.4 & {\bf 26.83}   & 26.32 & 26.26 & 26.07  &  26.50 &  26.08   & 25.26\\
\hline 0.5 & {\bf 28.51}   & 28.08 & 27.98 & 27.96   &  28.50  &27.72& 27.25\\
\hline
\end{tabular}
\label{Table:BarbaraPSNR}
\end{table*}


\subsubsection{VB and EM Inference}
\label{Sec:approxInfer}

The above results were based on an MCMC implementation. We next evaluate the accuracy and computation time of the different approximate algorithms.
The computation time of our approximate VB and EM (one iteration 2 sec at CSR$=0.4$) is very similar to TSW-CS (one iteration 1.62 sec).
The bottleneck in our methods is the evaluation of the Bessel function needed for inferences.
This can be improved ($e.g.$, by pre-computing a table of Bessel function evaluations).
Note, however, that our methods provide better results (Figure \ref{Fig:BarbaraPSNR} and Table \ref{Table:BarbaraPSNR}) than TSW-CS.

Table \ref{Table:BarbaraPSNR} also compares the different variational algorithms.
Recall that these involved solving an integral via Monte Carlo (MC), and we indicate the number of samples used by `num', giving ``VB(num)".
As the number of samples increases, the PSNR increases, but at the cost of computation time (the algorithm with $20$ MC samples takes 9.34 sec).
It is interesting to note that using the approximate VB and approximate EM, the PSNR of the reconstructed image is higher than TSW-CS (VB) and even VB(500).
Since the VB and EM provide similar results (VB is slightly better) and with almost the same computational time, we only show the results of a-VB.

\subsection{CS and block-DCT representation}

\begin{figure}[htbp]
  \centering
  \includegraphics[scale=0.2]{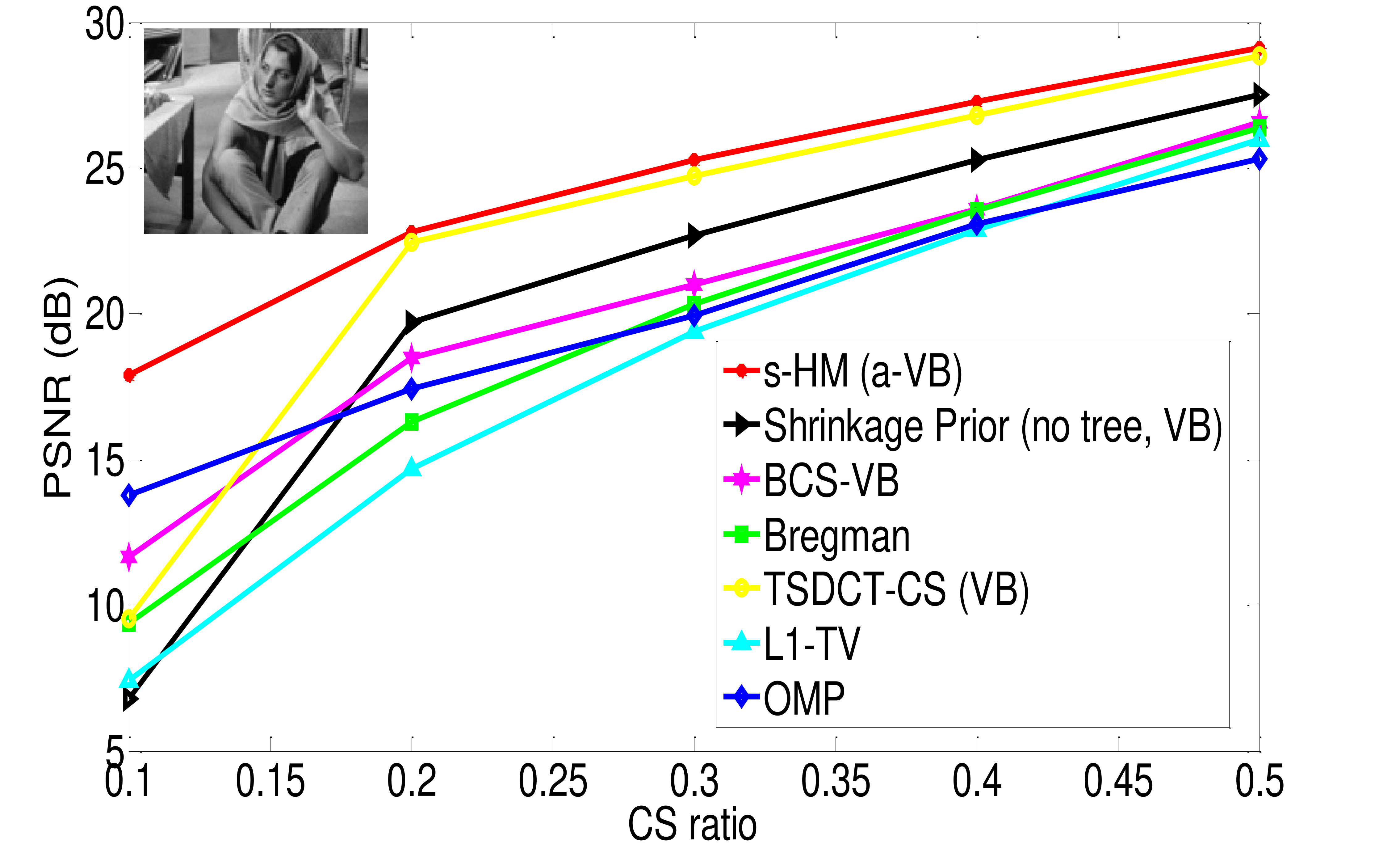}
  \caption{PSNR comparison of various algorithms for compressive sensing with block-DCT image representation. 
The ``TSDCT-CS" is the model implemented in \cite{He10SPL}, while the ``s-HM" denotes proposed model with block-DCT tree structure, and ``Shrinkage Prior (no tree)" symbolizes the flat model without tree-structure.}
    \label{Fig:BarbaraDCT}
\end{figure}

The principal focus of this paper has been on wavelet representations of images. However, we briefly demonstrate that the same model may be applied to other classes of tree structure. In particular, we consider an $8\times 8$ block-DCT image representation, which is consistent with the JPEG image-compression standard. Details on how this tree structure is constituted are discussed in \cite{Xiong96SPL}, and implementation details for CS are provided in \cite{He10SPL}).
Figure \ref{Fig:BarbaraDCT} shows CS results for ``Barbara" using each of the methods discussed above, but now using a block-DCT representation. We again note that the proposed hierarchical shrinkage method performs best in this example, particularly for a small number of measurements ($e.g.$, CSR of 0.1).

As in \cite{He10SPL}, for most natural images we have found that the wavelet representation yields better CS recovery than the block-DCT representation. The results in Figures \ref{Fig:BarbaraPSNR} and \ref{Fig:BarbaraDCT} are consistent with all results we have computed over a large range of the typical images used for such tests.

\begin{figure}[ht!]
        \centering
         \includegraphics[scale = 0.6]{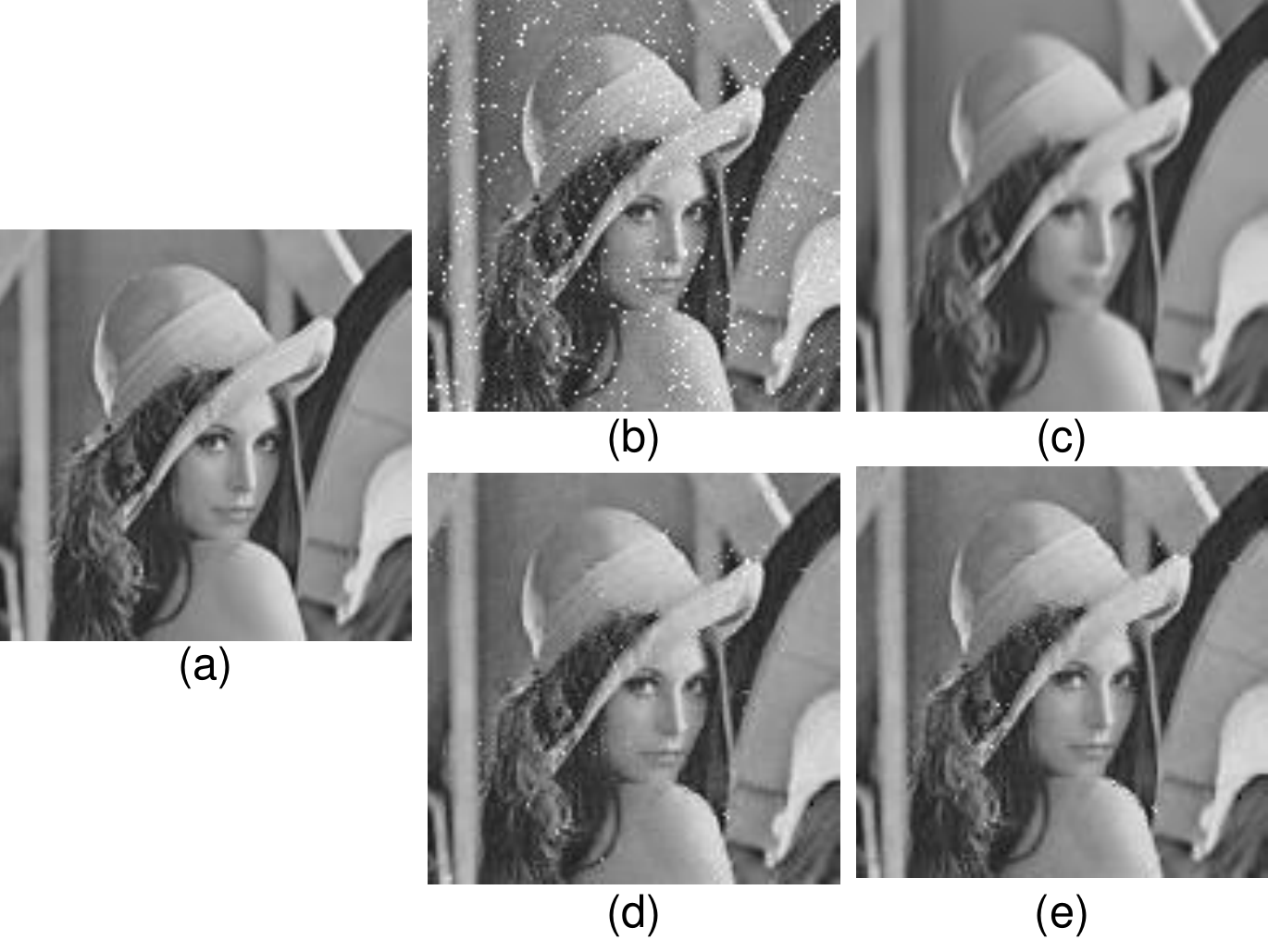}\\
         \caption{\small{Denoising. (a) original image, (b) noisy image (PSNR: 19.11dB), (c) dHBP (PSNR: 30.31dB), (d) shrinkage prior without tree (PSNR: 29.18dB), and (e) s-HM (PSNR: 30.40dB).}}
          \label{Fig:Lena_spiky}
 \end{figure}

\subsection{Denoising with spiky-plus-Gaussian noise}

We now apply the proposed model to denoising, with measurement noise that is a superposition of a Gaussian component and a spiky component. As in section \ref{Sec:Spiky},
we view this as a L\'evy
process, with the Brownian motion contributing the Gaussian part, and the Poisson jumps the spikes.
Inference is similar to Section \ref{Sec:inference}.
Figure \ref{Fig:Lena_spiky} compares the performance of our model with the state-of-the-art dHBP (dependent Hierarchical Beta Process) model in \cite{Zhou11AISTATS}, considering the same type of Gaussian+spiky noise considered there. Specifically, $3.2\%$ of the pixels are corrupted by spiky noise with amplitudes distributed uniformly at random over [0,1] (scaled images as before), and the zero-mean Gaussian noise is added with standard deviation 0.02.
While our results are similar to dHBP from the (imperfect) measure of PSNR, visually, the dHBP returns an image that appears to be more blurred than ours. For example, consider the detail in the eyes and hair in s-HM versus dHBP, Figures \ref{Fig:Lena_spiky}(c) and \ref{Fig:Lena_spiky}(e), respectively. Such detailed difference in local regions of the image are lost in the global PSNR measure.
Other commonly used images, including ``Barbara", ``Cameraman", ``Pepper" and ``House" were also tested, with similar observations.

In the dictionary-learning method of dHBP in \cite{Zhou11AISTATS}, neighboring patches are encouraged to share dictionary elements. This has the advantage of removing spiky noise, but it tends to remove high-frequency details (which often are not well shared between neighboring patches). In our work the spiky noise is removed because it is generally inconsistent with the wavelet-tree model developed here, that is characteristic of the wavelet coefficients of natural images but not of noise (Gaussian or spiky). Hence, our wavelet-tree model is encouraged to model the underling image, and the noise is attributed to the noise terms $\boldsymbol{w}$ and $\boldsymbol{n}$ in our model.

\section{Conclusions\label{sec:conclusions}}

We have developed a multiscale Bayesian shrinkage framework, accounting for the tree structure underlying a wavelet or block-DCT representation of a signal. The model that has been our focus is based on a gamma process hierarchical construction, with close connections to adaptive Lasso \cite{ZouAdaLasso}. 
Specially, we have proposed a tree-structured adaptive lasso.
This is but one class of L\'{e}vy process, and in \cite{Polso12Levy} it was demonstrated that many shrinkage priors may be placed within the context of L\'{e}vy processes. We have articulated how any of these shrinkage priors may be extended in a hierarchical construction to account for the tree-based representation of many bases. In that context, we have discussed how the Markov-based spike-slab hierarchical priors developed in \cite{He09SPT} may be viewed from the perspective of a L\'{e}vy process, with compound-Poisson L\'{e}vy measure. The proposed approach yields state-of-the-art results, using both a wavelet or block-DCT based image representation, for the problems of CS image recovery and denoising with Gaussian plus spiky noise. For the latter, we have made connections to dictionary learning, which is widely employed for such applications.

The proposed model accounts for the fact that wavelet and block-DCT coefficients are compressible, but not exactly sparse. By contrast, \cite{He09SPT} imposes exact sparsity, which tends to yield results that over-estimate the noise variance. This characteristic has manifested results from the proposed hierarchical shrinkage prior that are consistently better than that in \cite{He09SPT}, for CS signal recovery.

Concerning future research, in this paper we have considered only one class of L\'{e}vy process constructions, although the tree-based structure may be applied to any. It is of interest to examine other classes of L\'{e}vy processes, and applications for which they are particularly relevant. For example, in applications for which there are only a very few large coefficients, the symmetric alpha-stable L\'{e}vy process \cite{Wolp:Clyd:Tu:2011} may be most appropriate.

In this paper it has been assumed that the tree-based basis is known. In some applications one is interested in learning a tree-based dictionary matched to the data  \cite{Jenatton}. It is of interest to examine the utility of the proposed approach to this application, particularly because of its noted connection to dictionary learning. The proposed shrinkage priors have the advantage of, when viewed from the standpoint of the log posterior, being closely connected to many of the methods used in the optimization literature \cite{Jenatton}.

\bibliographystyle{IEEEtran}


\end{document}